\PassOptionsToPackage{hyphens}{url}
\documentclass[10pt, twocolumn]{amsart}

\usepackage{epstopdf}
\usepackage[caption=false]{subfig}

\usepackage[margin=2.5cm]{geometry}
\usepackage{enumitem}
\setlist{leftmargin=5.5mm}
\usepackage{caption}
\usepackage{multicol}
\usepackage{abstract}

\usepackage[round]{natbib}

\usepackage[T1]{fontenc}
\usepackage[utf8]{inputenc}
\usepackage{booktabs}

\usepackage{twemojis}

\usepackage[inkscapeformat=pdf]{svg}

\usepackage[colorlinks,bookmarksopen,bookmarksnumbered,citecolor=red,urlcolor=red]{hyperref}
\usepackage[nameinlink]{cleveref}
\theoremstyle{plain}

\theoremstyle{definition}

\theoremstyle{remark}

\title{Mapping the Technological Future: A Topic, Sentiment, and Emotion Analysis in Social Media Discourse}

\author{Alina Landowska}
\author{Maciej Skórski}
\author{Krzysztof Rajda}

\begin{document}


\twocolumn[

\maketitle

\begin{onecolabstract}
People worldwide are currently confronted with a number of technological challenges, which act as a potent source of uncertainty. The uncertainty arising from the volatility and unpredictability of technology (such as AI) and its potential consequences is widely discussed on social media. This study uses BERTopic modelling along with sentiment and emotion analysis on 1.5 million tweets from 2021 to 2023 to identify anticipated tech-driven futures and capture the emotions communicated by 400 key opinion leaders (KOLs). Findings indicate positive sentiment significantly outweighs negative, with a prevailing dominance of positive anticipatory emotions. Specifically, the 'Hope' score is approximately 10.33\% higher than the median 'Anxiety' score. KOLs emphasise 'Optimism' and benefits over 'Pessimism' and challenges. The study emphasises the important role KOLs play in shaping future visions through anticipatory discourse and emotional tone during times of technological uncertainty.
\end{onecolabstract}

\keywordsname{ Anticipatory Discourse, Anticipatory Emotions, Twitter, Discourse Analysis, Corpus Linguistics, Text Mining, Natural Language Processing, Topic Modelling}

]


\section{Introduction}
The future appears to be dominated by Artificial Intelligence (AI). Technology — particularly AI — exemplifies anticipation and uncertainty by replicating human intelligence. To the extent that we concern ourselves with the future, we oscillate between hope and anxiety. Hope and anxiety are recognised as uncertainty emotions \citep{gordon_emotions_1969, gordon_structure_1987}. On the other hand, broader anticipatory emotions include hope, trust, fear, and anxiety \citep{castelfranchi_anticipation_2011, feil_anticipatory_2022}.

Anticipation plays a critical role in decision-making and behaviour \citep{moore_data-frame_2011, grabenhorst_two_2021} as the desire to reduce uncertainty is a potent motivator of social behaviour \citep{hirsh_psychological_2012}, as people constantly seek to estimate and reduce uncertainties in social interactions to enhance their productivity, well-being, and ultimately their survival as social beings \citep{feldmanhall_resolving_2019}. The ability to navigate and adapt to uncertain futures becomes crucial. 

The discourse on social media is replete with envisaged future scenarios. These scenarios serve as a roadmap for users, guiding their expectations and informing their actions \citep{koivunen_anticipation_2023, tavory_coordinating_2013}. Influencers play a significant role in shaping public perceptions and the narrative surrounding these uncertainties \citep{acerbi_storytelling_2022, acerbi_cultural_2019, jin_following_2014, landowska_what_2023, spry_celebrity_2011, tur_effect_2022}. For instance, their anticipation of AI and automation, while often seen as progress, also brings up concerns about job loss, privacy, and tech power concentration. This study not only uncover the anticipatory discourse on the X platform but also scrutinizes the power dynamics in tech discourse, particularly the impact of key opinion leaders (KOLs) in moulding public views of tech futures. Based on a corpus of anticipatory discourse \Cref{tab:tweets_examples}) from Twitter spanning 2021 to 2023, this study integrates BERTopic modelling, sentiment, and emotion analyses to explore the futures represented by technology influencers on social media and to assess the emotional impact of these anticipated visions articulated to the public audiences.

Research on how individuals anticipate and structure future events has been central in social theory, e.g., \citep{adam_time_1990, bergmann_problem_1992, emirbayer_what_1998, abbott_time_2001, mische_projects_2009}. Every human interaction inherently involves a connection to future events \cite{tavory_coordinating_2013}. When individuals engage with each other, they orient themselves towards future circumstances, and consequently it becomes imperative for individuals to simultaneously navigate the inherent uncertainties associated with these future-oriented interactions(\textit{ibid.}, p.~909). 

\begin{table}
    \centering
    \begin{tabular}{p{1cm} p{6cm}}
    \toprule
    Topic & Futurist's anticipatory tweet\\
    \midrule
     9 & ''Next level of \#remotework - will working from home (\#WFH) be the new standard for the \#futureofwork?'\\
     40 & 'Will you learn to trust artificial intelligence next year?'\\
     54 & 'How is Artificial Intelligence Revolutionizing the Educational Sector?'\\
     78 & '\#Who’s Next? Security Guards? Yet another job that will be replaced by robots?'\\ 
     90 & 'When will we see the first \#AI generated hit song? Book? \twemoji{thinking face}\\
    \bottomrule
    \end{tabular}
    \caption{Examples of Anticipatory Discourse of Future Influencers in the Analysed Corpora}
    \label{tab:tweets_examples}
\end{table}

By employing a triangulated methodological approach (e.g., \citep{egbert_using_2019}, we aim to integrate three distinct methods for discourse analysis of a tweet-based corpus: BERTopic modelling (contextual corpus analysis), Keywords analysis (non-contextual corpus analysis), and lexicon-based analysis (sentiment, emotion, attitude analyses). Our goal is to identify the leading technologies and KOLs, how technological topics (anticipated futures) evolve and interrelate over time, and the sentiments, anticipatory emotions, and attitudes associated with them within a corpus of approximately 1.5 million tweets spanning the years 2021 to 2023. Through this combination, we seek to address the following research questions:

\begin{itemize}
\item \textbf{[Rq1]} What specific technology-driven futures are being anticipated by KOLs on platform X and how do they evolve over time and interrelate with each other as well as to what extent are these anticipated futures aligned with the concepts of 'present future' and 'future present'?
\item \textbf{[Rq2]} Who are the Key Opinion Leaders (KOLs) driving these discussions, and how does their influence shape the prominence of specific technologies on platform X?
\item \textbf{[Rq3]} What are the specific sentiments (either positive or negative), anticipatory emotions, and attitudes (either optimistic or pessimistic) expressed by KOLs towards specific technology-oriented futures?
\end{itemize}

\section{Literature Review}

\textbf{Anticipating (Abstract or Concrete) Futures.} We recognise two main categories of the future: a. ‘present futures’ \citep{luhmann_differentiation_1982} are ‘pre-given futures’ rooted in the past, i.e., lived ones \citep{adam_futures_2011}; and b. ‘future present’ \citep{luhmann_differentiation_1982} are ‘futures-in-the-making’ that are possibly latent, growing, and changing ones, i.e., living ones \citep{adam_futures_2011}. The 'present future' refers to the concrete, i.e. practical, presentational, explicit, \citep{ poli_introduction_2017, poli_anticipation_2014} and it is the contextual, embodied, and embedded future \citep{adam_future_2007}; and the 'future present' refers to the abstract (symbolic, representational, implicit) future that lacks context and content, creating an opportunity for exploration and processing \citep{adam_future_2007}.
'Present futures' are linear continuations of the past in the present \citep{poli_anticipation_2014}, which shape the future by present means \citep{adam_future_2007, miller_futures_2007}. The value of these closed-type futures is calculated against its alternatives (forecasts) and traded as commodities \citep{adam_future_2007}; the future with the highest value is the one with the largest profit. Because of the optimistic or pessimistic character of their transformed or reinterpreted visions, they serve as projections aiming at evoking hope or fear \citep{luhmann_differentiation_1982}.

'Future presents' are latent, but they can be recognised and foreseen, thus impacting the present by entering into it and being used in the present \citep{adam_futures_2011}. These futures are conceptualised, isolated from their contexts, and technologically prejudiced \citep{luhmann_differentiation_1982}, which makes them limitless possibilities, leading to high uncertainty \citep{adam_future_2007, beckert_capitalism_2013, poli_anticipation_2014, poli_introduction_2017}. These abstract futures are open-type futures with abstract values that are freely traded on the assumption that they can be computed at any location and at any time, and it may be used for any situation \citep{adam_future_2007}. The 'future presents' are unpredictable and evoke fictional expectations based on the “as if” rule \citep{beckert_capitalism_2013}. Future presents seems to be anticipatory in their character because of their ‘use-of-the-future’ orientation \citep{poli_introduction_2017}.

\textbf{Anticipatory Emotions.} Anticipation plays a crucial role in the generation of emotions, and the literature distinguishes between future-oriented emotions for anticipatory and anticipated experiences. Anticipatory emotions encompass present feelings related to an upcoming event \citep{bagozzi_goal-directed_1998, baumgartner_future-oriented_2008, feil_anticipatory_2022}, whereas anticipated emotions pertain to the expected emotional responses a person may have when faced with a future occurrence \citep{baumgartner_future-oriented_2008, feil_anticipatory_2022, perugini_role_2001}. Anticipatory emotions primarily manifest as prospective feelings of hope and fear and anticipated emotions tend to be retrospective, including sensations like relief, satisfaction, disappointment, and anger \citep{baumgartner_future-oriented_2008, feil_anticipatory_2022}. Hope responds to possible future outcomes, while fear is associated with negative ones \citep{castelfranchi_anticipation_2011, macleod_anticipatory_2017, vazard_feeling_2024}. Unmet hope is leading to disappointment, whereas the absence of a feared event brings relief \citep{baumgartner_future-oriented_2008, mowrer_learning_1960}.

Moreover, \citep{plutchik_general_1980} identifies anticipation as one of the eight primary bipolar emotions. At primary dyads, it interacts with joy to create optimism, and with anger to result in aggressiveness \citep{plutchik_nature_2001}. Secondary dyads, two petals apart, pair anticipation with trust, leading to hope, and with disgust, culminating in cynicism (\textit{ibidem}). Tertiary dyads, three petals apart, combine anticipation with sadness, resulting in pessimism, and with fear, leading to anxiety (\textit{ibidem}). In this emotional spectrum, anticipation stands in contrast to surprise (\textit{ibidem}).

\section{Data and Methods}\label{sec:methodology}

\textbf{Data Collection.} The dataset was collected using the scraping library \texttt{snsscrape} ~\citep{justanotherarchivist_snscrape_2023}. Tweets were sourced from about 400 technology influencers' feeds published in the timeframe from January 1, 2021, to March 31, 2023. The selection of KOLs was meticulously carried out by our team of experts. \Cref{tab:dataset_stats} provides a comprehensive overview of the size and diversity of our dataset.

\begin{table}[]
    \centering
    \begin{tabular}{ l l  }
    \toprule
     Unique accounts & 400 \\ 
     Unique texts & 1,200,003 \\  
     Unique tweets (timestamp+text) & 1,458,018 \\  
    \bottomrule
    \end{tabular}
    \caption{Dataset summary statistics}
    \label{tab:dataset_stats}
\end{table}

\textbf{Data Cleaning.} To ensure the accuracy of this analysis, we preprocessed the tweets by removing URLs, emails, and user handles, which are identifiers for X users. 

\textbf{Data Analysis.} We employed a comprehensive suite of Natural Language Processing techniques to conduct an in-depth analysis of the data we collected. Appropriate non-parametric statistical tests were used to assess significance of findings.

\textit{Contextual Corpus Analysis.} Following prior work on extracting topics from Twitter data \citep{bogdanowicz_dynamic_2022, egger_topic_2022, landowska_what_2023, yang_standing_2021}, we trained the state-of-the-art BERTopic model implemented in the \texttt{BERTopic} library~\citep{grootendorst2022bertopic}. As the default setting of BERTopic is vulnerable on data volume, to perform analysis efficiently, we used an online implementation of the algorithm using: \textit{IncrementalPCA} dimensionality reduction model, \textit{MiniBatchKMeans} clustering algorithm and a \textit{OnlineCountVectorizer} vectorisation scheme, with an \textit{all-MiniLM-L6-v2} embeddings base applied to raw tweets. BERTopic analysed the dataset and categorised it into 100 contextually coherent, distinct topics. We visualised the topics using the method of \citep{sievert_ldavis_2014}, implemented in the library \texttt{pyLDAvis}~\citep{mabey_pyldavis_2021}.

\textit{Non-contextual Corpus Analysis.} Our analysis of the non-contextual corpus consists of identifying KOLs and keywords. KOLs in tech discussions were identified based on their tweet volume on the X platform \citep{casalo_influencers_2020}. Indicating keywords required utilising a compact English model from the  Spacy\footnote{\url{https://spacy.io/}} package. This model analyses the given corpus, assigning respective metadata to each sentence and word, ultimately generating an annotated corpus. The textual data underwent processing through lemmatisation and Part-of-Speech (POS) tagging. The lemmatised form of any word serves as a keyword, excluding stop words. Subsequently, we computed the global attributes of the corpus, including the most distinctive keywords.

\textit{Sentiment and Emotions Analysis.} To analyse sentiment, we leveraged the Brand24 model \citep{augustyniak_massively_2023, rajda_assessment_2022}. This recent model is based on a fine-tuned multilingual LLM and is specifically tailored for analysing social media data. The Brand24 sentiment classifier operates in a three polarities (positive, negative, and neutral). The average sentiment score was defined as follows:
\begin{align}\label{eq:sentiment}
\mathrm{Score}=\frac{N^{+}-N^{-}}{N^{-}+N^0+N^{+}}
\end{align}
where $N^{-},N^0$ and $N^{+}$ denote the number of tweets classified as positive, neutral, and negative, respectively. For analysing emotions, we chose to adopt the enhanced and expanded model introduced by \citep{mohammad_semeval-2018_2018}, which is accessible as a functional framework shared by \citep{camacho-collados_tweetnlp_2022}. We utilised the official implementation available from the HuggingFace repository\footnote{\url{https://huggingface.co/cardiffnlp/twitter-roberta-base-emotion-multilabel-latest}}. The model classifies emotions in a multi-label schema, meaning that each text is assigned a separate score for each of the 11 emotions (sadness, disgust, anger, fear, anticipation, surprise, joy, love, trust, pessimism, optimism). Pessimism and optimism scores were rejected. Following Plutchik \cite{plutchik_general_1980}, to define scores of hope and anxiety, for each topic, we paired anticipation with trust to obtain hope scores, while also linking anticipation with fear, resulting in anxiety (\textcolor{blue}{please refer to \Cref{eq:sentiment}}).

\begin{align}\label{eq:secondary_emotions}
\small
\begin{aligned}
    \mathrm{Score(Anxiety)} & = \mathrm{Score(Anticipation)} + \mathrm{Score(Fear)} \\
    \mathrm{Score(Hope)} & = \mathrm{Score(Anticipation)} + \mathrm{Score(Trust)},
\end{aligned}
\end{align}

and the topics are scored by averaging, respectively. We incorporated an emotion colour palette inspired by Plutchik’s wheel of emotions from ~\citep{semeraro_pyplutchik_2021}. This visually enhanced our findings and provide a more intuitive representation of emotional states. Despite the frequent use of the terms optimism and pessimism, there is a noticeable lack of consensus on their precise definitions. In this study, we propose a coherent approach by interpreting them as a manifestation of a ‘can-do’ attitude \citep{bortolotti_optimism_2018, jefferson_what_2017}. We suggest, according to \citep{plutchik_nature_2001} and \citep{tenhouten_emotions_2023}, to define optimism as the cumulative score of anticipation and joy, while pessimism as the combined score of anticipation and sadness. This approach provides a more structured and verified method to measure these attitudes as follows:
\begin{align}\label{eq:secondary_attitude}
\small
\begin{aligned}
\mathrm{Score (Optimissism)} &= \mathrm{Score(Anticipation)}+ \mathrm{Score(Joy)}\\
\mathrm{Score (Pessimism)} & = \mathrm{Score(Anticipation)}+ \mathrm{Score(Sadness)} 
\end{aligned}
\end{align}

\section{Results}

\subsection{Technology-Driven Topics}

\textbf{Distinct Topics}. After several rounds of training the BERTopic model, we determined that the optimal number of distinct topics was 100. Each topic is linked to a specific set of tweets, enabling a deeper understanding of the principal themes within the dataset. For a comprehensive list of all topics (please refer to the supplementary repository). The Coherence Value measure (please see \citep{roder_exploring_2015}) for 25 topics is about 66\%, demonstrating that topics are of high-quality (as evidenced by other studies on Twitter/X data, under this metric, BERTopic rarely achieves scores higher than 70\% \citep{campagnolo_topic_2022,santakij_analyzing_2024,austin_uncovering_2024} unless for small and coherent corpora \citep{chen_leveraging_2023}).

\begin{figure}
    \centering
    \includegraphics[width=0.99\columnwidth]{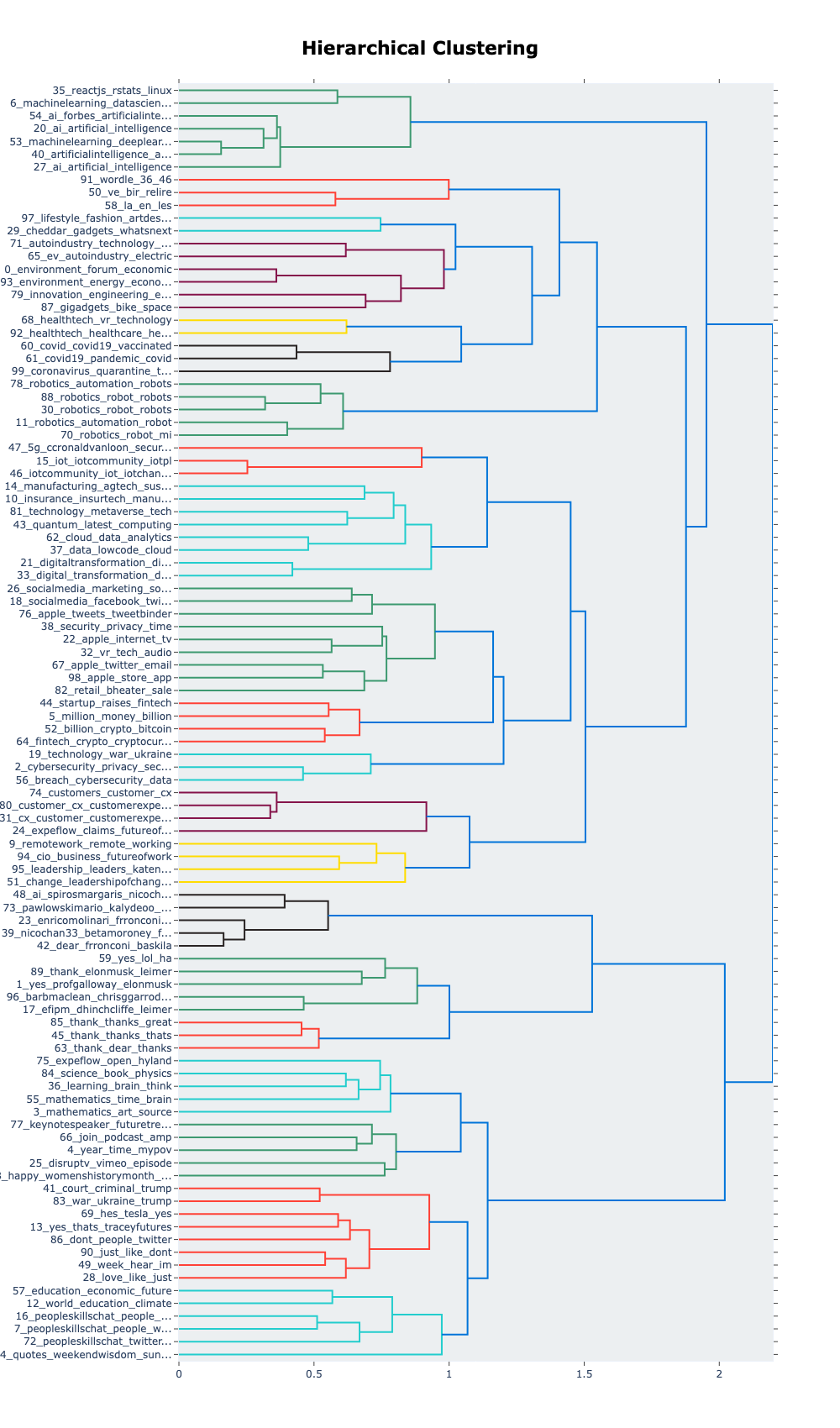}
    \caption{The Hierarchical Clustering of Anticipated Technology-Driven Futures}
    \label{fig:clusters}
\end{figure}

\textbf{Hierarchical Clustering.} BERTopic model also generated 21 unique clusters from 100 topics (please refer to \Cref{tab:BERT_clusters}). The relationships between the documents and these themes are elucidated, as depicted in the hierarchical clustering dendrogram (please refer to \Cref{fig:clusters}). These groups show how different tech themes like AI, ML, and Data Science are influencing each other as they develop within the discussing community. The dendrogram illustrates how topics are related to each other based on their distance in the high-dimensional space, where topics are closer if they are more similar (please refer to Intertopic Distance Map in the supplementary repository).

\textbf{Engagement in discussions on futures.} The range of tweet counts across topics is 23,144 tweets (from 7,633 to 30,777). The average engagement of discussion (the mean count of tweets per topic) is 14,580 tweets. Approximately 41.75\% of the topics exceed this mean count. After excluding non-technology related discussions like 'Time Reflections' or 'Gratitude,' the top 5 tech-driven topics include Topic 25 (27,231 tweets), Topic 0 (27,103 tweets), Topic 83 (22,485 tweets), Topic 6 (21,827 tweets), and Topic 46 (21,635 tweets). These topics are discussed significantly more frequently than the bottom 5 below the median, such as Topic 70 (7,633 tweets), Topic 99 (8,418 tweets), Topic 68 (8,430 tweets), Topic 79 (8,792 tweets), and Topic 14 (9,096 tweets).

\textbf{Futures Evolution.} Some topics exhibit seasonal patterns, characterised by regular peaks and troughs, while others demonstrate clear trends—either increasing or decreasing over time (refer to \Cref{fig:topic-evolution}). Time trends for each topic can be accessed in the supplementary repository. Certain topics consistently reached all-time highs over the examined time period compared to other topics. For instance, Topic 53 shows sustained interest and growth, reflecting its evolving role in shaping future technological landscapes. In contrast, other topics demonstrate fluctuating patterns, suggesting shifts in interest and relevance over time. Certain topics demonstrate sharp increases or decreases at specific points. For instance, Topic 0 (the blue line) reached an all-time high with two significant peaks. The first peak occurred during the WEF from 17 to 21 January 2022, and the second during the 27th UN COP 27 held from 6 November until 20 November 2022. Topic 46 (the red line) showed a peak around of the Special Annual Meeting of the WEF from May 25-28, 2021. Some topics, such as 83 (the violet line), manifest notable fluctuations, while others like Topic 78 (the pink line) remain relatively stable until they drop off at the end. All topics demonstrate a consistent presence over time, with none of them showing a drop-off, indicating their sustained relevance. Finally, there are clear intertopic relations over time. For instance, Topics 27 (the light green line), 32 (the dark green line), and 54 (the orange line) resonate with each other.

\begin{table}[]
    \centering
    \begin{tabular}{p{0.8cm} p{2cm} p{4.2cm}}
    \toprule
    Cluster & Topics & Leitmotiv\\
    \midrule
    1 & 35, 6, 54, 20, 53, 40, 27 & AI, ML, DL, Big Data, NLP\\ 
    2 & 91, 50, 58 & Entertainment and Gaming\\  
    3 & 97, 29 & Fashion and Lifestyle\\ 
    4 & 71, 65, 0, 93, 79, 87 & AV, Automatic Industry, EV, Eco-energy, Climate Change, Innovative engineering, Drones, Space \\ 
    5 & 68, 92 & Healthcare, AI\\ 
    6 & 60, 61, 99 & Covid-19 and Quarantine\\ 
    7 & 78, 88, 30, 11, 70 & Automation, Robots Engineering, Robotics, Autonomous Robots, AI\\ 
    8 & 47, 15, 46 & IoT, 5G, Security, Digital Transformation, Industry 4.0, Edge Computing\\
    9 & 14, 10, 81, 43, 62, 37, 21, 33 & Quantum Computing, Data Analytics, Digital Transformation, Tech Innovations,  Metaverse\\ 
    10 & 26, 18, 76, 38, 22, 32, 67, 98, 82 & Women in Tech, Social Media, Payments, Retail, Privacy, Security\\
    11 & 44, 5, 52, 64 & Start-up, Fintech, Cyrptocurrency\\
    12 &  19, 2, 56 & Cybersecurity, War, Ukraine\\
    13 &  74, 80, 31, 24 & CX, UX, Future of Claims, Insurtech\\
    14 &  9, 94, 95, 51 & Future of Work, Future of Leaders, Leadership of Change, Strategy, Management\\
    15 &  48, 73, 23, 39, 42 & AI, Data Science, AV, VR, Digital Transformation, Metaverse\\
    16 &  59, 89, 1, 96, 17 & Gratitude (Thanks, Acknowledgements\\
    17 &  85, 45, 63 & Appreciation (Thanks, Great)\\
    18 &  75, 84, 36, 55, 3 & Mathematics, Life, Questioning\\ 
    19 &  77, 66, 4, 25, 8 & Time Reflections (Year, Mypov, Women in History)\\ 
    20 &  41, 83, 69, 13, 86, 90, 49, 28 & Politics, Policies, Ukraine War, Social Media\\ 
    21 &  57, 12, 16, 7, 72, 34 & Future, Economics, Education, People Skills, Collaborative Enterprise, Climate, Teamwork\\ 
    \bottomrule
    \end{tabular}
    \caption{The organisation of Topics into Clusters by BERTopic Analysis}
    \label{tab:BERT_clusters}
\end{table}

\begin{figure}
\centering
\includegraphics[width=0.99\columnwidth]{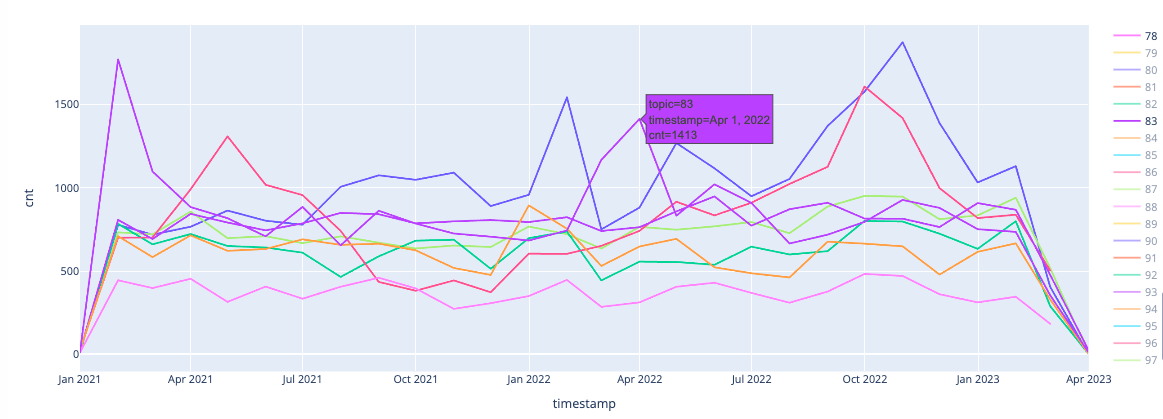}
\caption{The Dynamics of Selected Topics Over Time}
\label{fig:topic-evolution}
\end{figure}

\subsection{KOLs and Trending Technologies}

\begin{figure}
    \centering
    \includesvg[width=0.99\linewidth, inkscapelatex=false,]{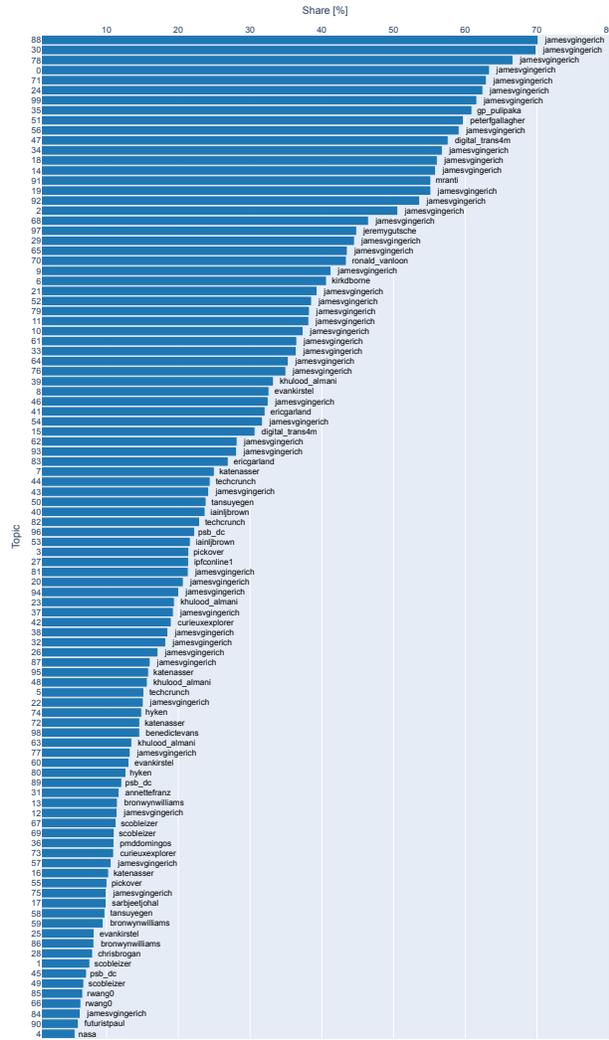}
\caption{Topic leaders.}
\label{fig:topic_leaders}
\end{figure}

\begin{figure}
    \centering
    \includegraphics[width=0.99\linewidth]{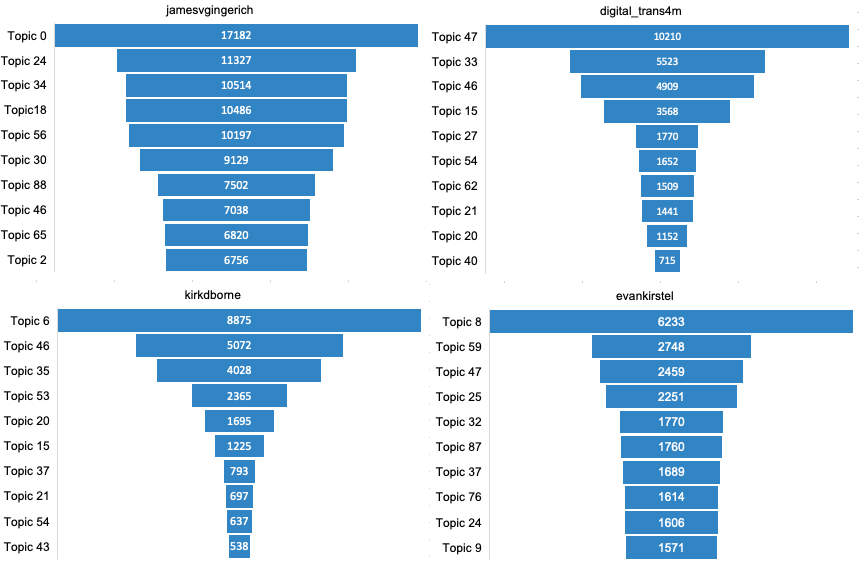}
    \caption{Number of Tweets by Top-Performing KOLs in the Corpus}
    \label{fig:selected_performers}
\end{figure}

\begin{figure}
    \centering
    \includegraphics[width=0.99\linewidth]{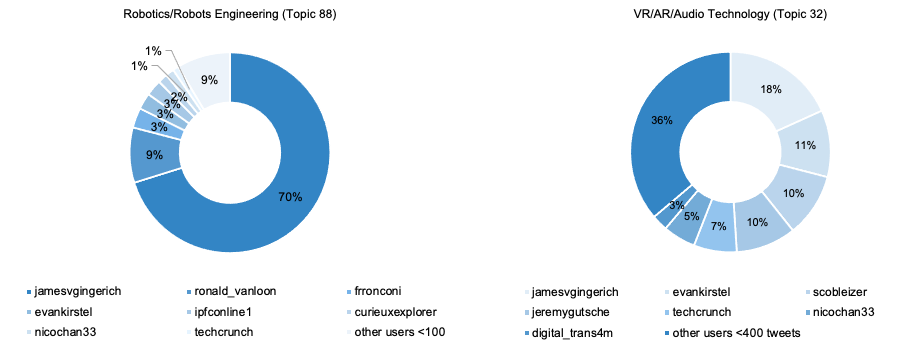}
    \caption{Contribution Distribution Across Topics in Robotics/Robots Engineering versus VR/AR/Audio Technology}
    \label{fig:KOL_2_topics_88_versus_32}
\end{figure}

\textbf{Leading performers.} KOLs are subject matter experts who make significant contributions to specific topics and the prominence of specific technologies on platform X. A small number of them dominate the discourse, with the top 10 KOLs generating 13\% (187,645 tweets) of the total tweets (refer to \Cref{fig:topic_leaders}). They play a pivotal role in steering discussions across 22 topics, dominating these conversations with a substantial share of over 49\%. The data clearly shows distinct specialisation among the KOLs, except for jamesvgingerich, who is active in 46 topics. \Cref{fig:selected_performers} showcases the top performers jamesvgingerich is the standout performer, leading in all 10 topics. Among the other key performers, digital\_trans4m is the leading contributor in Topic 47, while kirkdborne leads in Topic 6, and evankirstel leads in Topic 8. However, these users do not hold leading positions in other topics. While most topics (e.g., Topic 32) are evenly distributed among users, Topic 88 stands out as an exception (please refer to \Cref{fig:KOL_2_topics_88_versus_32}). Here, user jamesvgingerich has a significant lead responsible for a staggering 70\% of all tweets on this topic, amounting to 7502 tweets. In comparison, the second-highest contributor, ronald\_vanloon, has made 951 tweets, which constitutes 9\% of the total contributions to this topic. 

\begin{figure}
    \centering
    \includegraphics[width=1.0\linewidth]{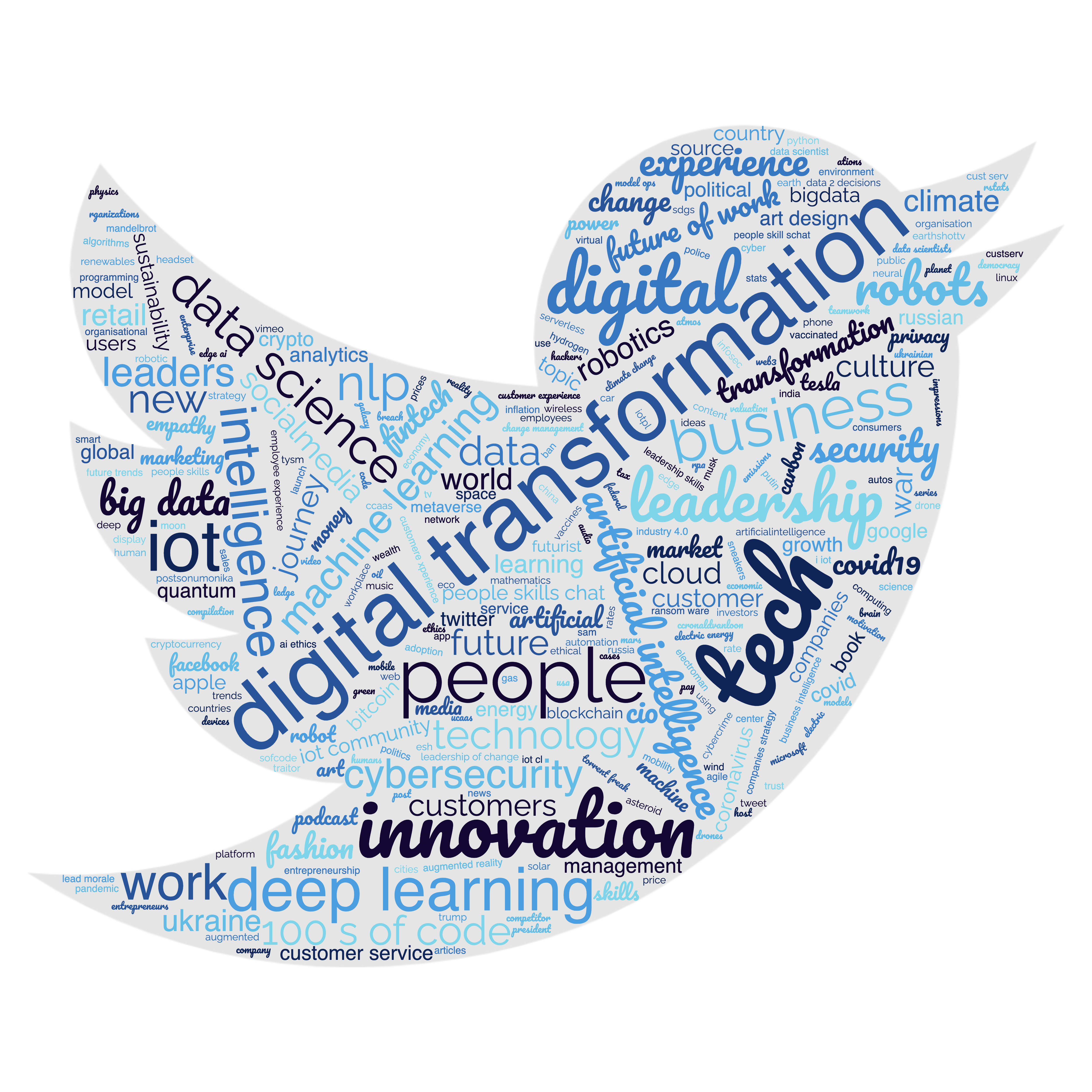}
    \caption{The Most Frequently Occurring Keywords}
    \label{fig:keywords}
\end{figure}

\begin{figure}
    \centering
\includesvg[width=0.99\linewidth,inkscapelatex=false]{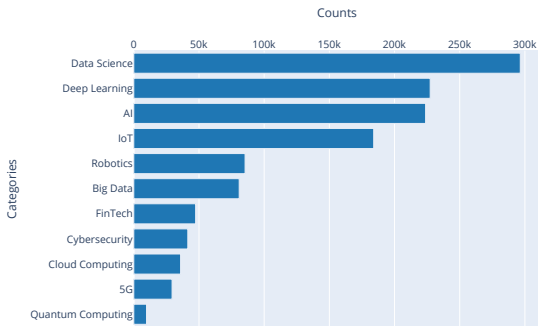}
    \caption{Leading Technologies}
    \label{fig:top_technologies}
\end{figure}

\textbf{Leading technologies.} We also identified the top 100 keywords (the most discussed in the dataset) based on their frequency of occurrence in the entire dataset (refer to \Cref{fig:keywords}). \Cref{tab:repres_keywords} showcases the most representative keywords for 5 randomly selected topics. Every keyword, each representing a distinct topic, can be found in the (name removed for reviewing) repository. Keywords, although frequently oversimplified markers, lack context and nuance. However, through analysis, they enable the identification of leading technologies extensively discussed among KOLs (refer to \Cref{fig:top_technologies}).

\begin{table}[]
    \centering
    \begin{tabular}{p{1cm} p{6cm}}
    \toprule
    Topic & Related keywords\\
    \midrule
    2 & 'cybersecurity', 'privacy', 'security', 'cyber', 'cyberattack', 'cyberwar', 'cybercrime', 'hackers', 'data', 'ukraine'\\ 
    11 & 'robotics', 'automation', 'robot', 'ai', 'artificialintelligence', 'robots', 'robotic'\\  
    40 & 'artificialintelligence', 'ai', 'intelligence', 'artificial', 'nlp', 'machinelearning', 'robots', 'deeplearning', 'read', 'bigdata' \\ 
    43 & 'quantum', 'latest', 'computing', 'trends', 'cloud', 'quantumcomputing', 'tech' \\ 
    79 & 'innovation', 'engineering', 'energy', 'technology', 'scientists', 'interesting', 'future', 'space', 'science', 'world'\\ 
    \bottomrule
    \end{tabular}
    \caption{The Most Representative Keywords in Selected Topics}
    \label{tab:repres_keywords}
\end{table}


\subsection{Sentiment and Emotions Across Topics and Clusters}

\begin{figure}
    \centering
\includesvg[width=1.0\linewidth,inkscapelatex=false]{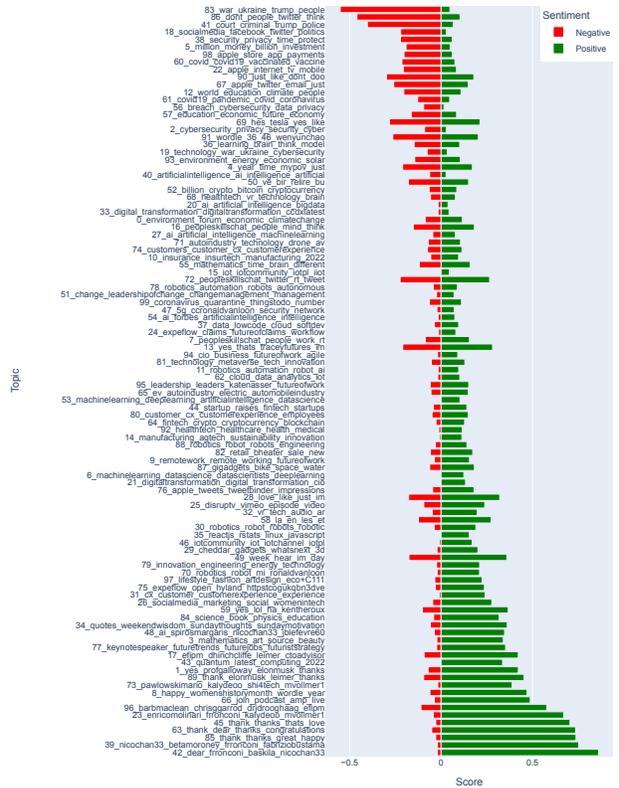}
\caption{Comparative Analysis of Positive and Negative Sentiment Distribution Across Various Topics}
\label{fig:sentiment_clusters}
\end{figure}

\textbf{Sentiment Score.} The sentiment analysis of the dataset reveals that a significant majority, i.e. 68.9\% of the tweets are classified as neutral, followed by positive (21.5\%) and then negative (9.6\%) (please refer to \Cref{fig:sentiment_clusters}). The most neutral scores associated to Topics 0, 6 25, 46, 40, 33, 27, 24, 47, 20, 44. The most positive sentiment dominates in Topics such as 59, 89, 1, 96, 17 85, 45, 63. The most negative sentiment is associated with Topic 83, 41, 18, 86, 2, and 56. The average sentiment score ranges from $-$0.5038 (very negative) to 0.8419 (very positive), with a mean of approximately 0.116, indicating an overall positive sentiment across all topics. This is evidenced by the statistically significant result of the one-sided Wilcoxon signed-rank test ($p=2\cdot 10^{-18})$ for comparison of positive and negative sentiment across topics.

\begin{figure}
    \centering
\includesvg[width=1.0\linewidth,inkscapelatex=false]{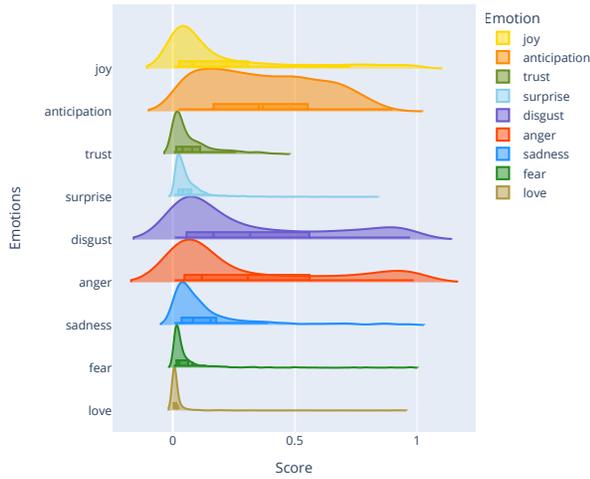}
\caption{Distributions of Emotions Across the Corpus}
\label{fig:emotions_descriptive}
\end{figure}

\begin{figure}[h!]
    \centering
\includesvg[width=1.0\linewidth,inkscapelatex=false]{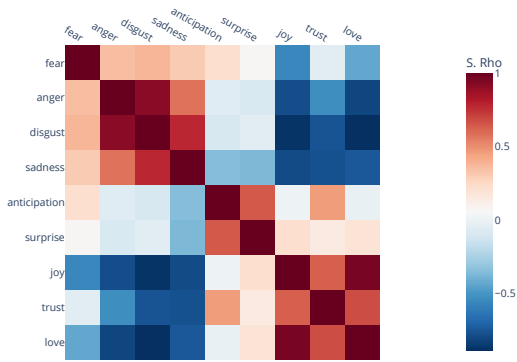}
\caption{Spearman's Rho Correlation of Emotions Across the Corpus}
\label{fig:emotions_corr}
\end{figure}

\begin{figure}[h!]
    \centering
\includesvg[width=0.99\columnwidth,inkscapelatex=false,]{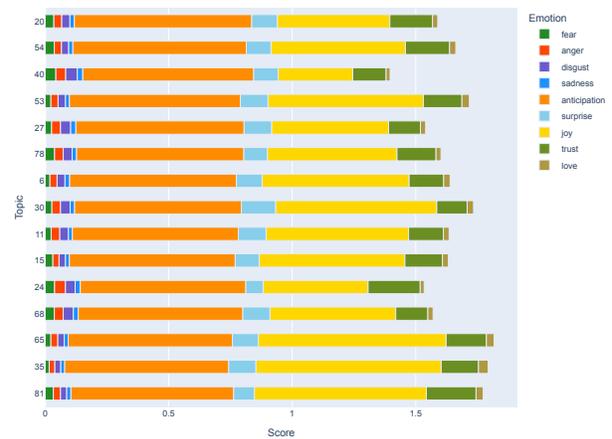}
    \caption{Emotions in Top-Anticipating Topics}
    \label{fig:top15_anticipation}
\end{figure}

\textbf{Emotions.} The distribution of emotions is presented in \Cref{fig:emotions_descriptive}, for more numerical details please refer to the repository. The 5 topics with the highest emotion scores (above the median) and the 5 topics with the lowest emotion scores (below the median) please refer to \Cref{tab:emotion_scores}. For top  anticipation scores please refer to \Cref{fig:top15_anticipation}. 
The statistical correlation analysis was performed with the use of non-parametric Spearman's rank correlation coefficient, demonstrating significance of correlation presence in all emotion pairs  ($p=10^{-16}$ for joint significance after Benferoni correction). The results are  summarised in~\Cref{fig:emotions_corr} and reveal that: i. the correlation (0.455) between anticipation and trust indicates that anticipation enhances trust; ii. With a moderate negative correlation (-0.353), that higher anticipation is associated with lower sadness; iii. the weak positive correlation (0.217) b implies that as anticipation increases, fear also tends to increase slightly; iv. the very weak correlation (0.007) suggests that anticipation and joy are largely independent. When anticipation increases, joy does not necessarily increase or decrease significantly. 


\begin{table}[]
    \centering
    \begin{tabular}{l l l}
     \toprule
     Emotion & \multicolumn{2}{c}{Topics} \\ & Highest Score & Lowest Score \\
    \midrule
     Joy & 34, 94, 14, 42, 31 & 52, 99, 55, 72, 13\\  
     Anticipation & 94, 14, 31, 77, 43 & 99, 55, 72, 13, 4 \\ 
     Trust & 51, 34, 94, 7, 80 & 72, 71, 47, 13, 4\\
     Disgust & 7, 80, 95, 0, 24 & 65, 78, 48, 39, 20\\
     Suprise & 77, 43, 79, 81, 84 & 17, 10, 96, 93, 49\\
     Anger & 34, 7, 80, 95, 0 & 78, 48, 39, 20, 62\\
     Sadness & 34, 7, 80, 95, 0 & 52, 1, 82, 99, 48\\
     Fear & 7, 95, 0, 81, 92 & 17, 96, 29, 37, 35\\
     Love & 34, 7, 80, 95, 14 & 68, 36, 19, 27, 40\\
    \bottomrule
    \end{tabular}
    \caption{Highest and Lowest Emotion Scores in Topics}
    \label{tab:emotion_scores}
\end{table}

\begin{figure}[h!]
    \centering
\includesvg[width=0.99\columnwidth,inkscapelatex=false]{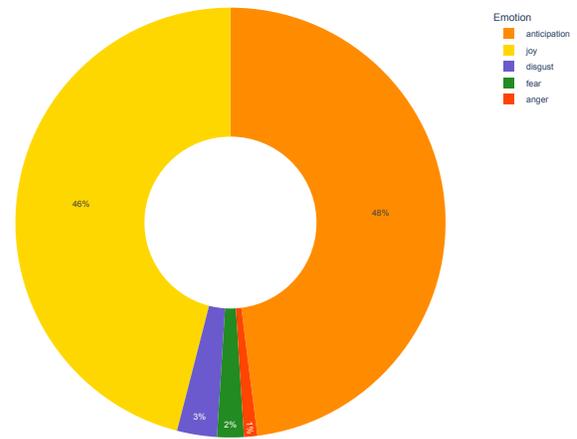}
    \caption{Dominance of Emotions Across All Topics}
    \label{fig:Emotion_in_percentages}
\end{figure}

At aggregated level, the data indicates a strong dominance of 'Anticipation' (48\%) and 'Joy' (46\%) across all topics (please refer to \Cref{fig:Emotion_in_percentages}). The remaining 6\% of topics were primarily characterised by negative emotions, i.e. 'Disgust' (3\%), 'Fear' (2\%), and 'Anger' at a minimal level of 1\%.

\begin{figure}
    \centering
    \includegraphics[width=0.99\columnwidth]{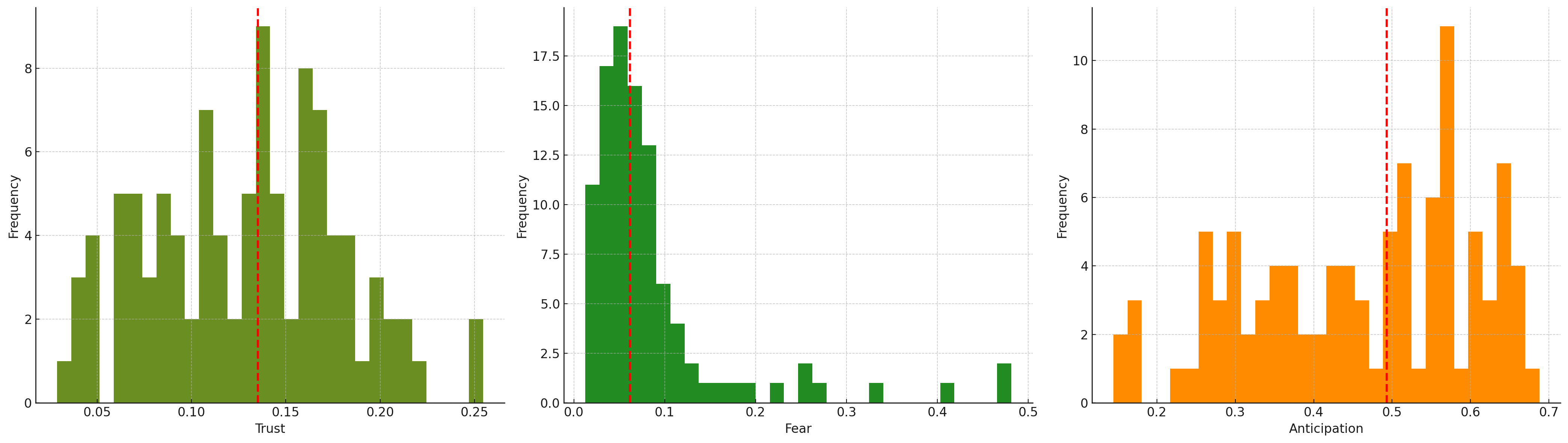}
    \caption{Distribution of Anticipation, Trust, and Fear in Topics}
    \label{fig:distribution_of_joy_trust_fear}
\end{figure}

\textbf{Anticipatory Emotions.} 'Anticipation’ correlates with both ‘Trust’ (leading to 'Hope') and ‘Fear’ (leading to 'Anxiety'), for the frequency of these emotion scores across topics please refer to \Cref{fig:distribution_of_joy_trust_fear}. The median 'Hope' score (0.6345) is approximately 10.33\% higher than the median 'Anxiety' score (0.5751). The highest 'Hope' scores were found in topics, e.g. 51, 94, 80, 14, 95, topics with the lowest 'Hope' scores were, e.g., 7, 0, 57, 9, 60. The highest 'Anxiety' scores were observed in topics, e.g. 94, 14, 31, 43, 79, whereas the lowest scores included, e.g. 51, 7, 80, 95, 0. The non-parametric one-sided Wilcoxon signed-rank test shows the 'Hope' dominates over 'Anxiety' across topics ($p = 5.58 \times 10^{-7}$).

\begin{figure}
    \centering
    \includesvg[width=0.99\columnwidth,inkscapelatex=false]{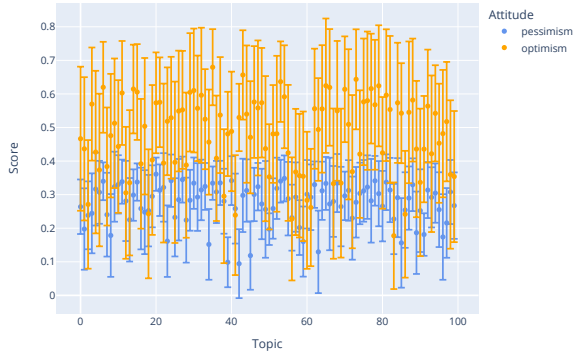}
    \caption{Distribution of Optimism and Pessimism Across All Topics}
    \label{fig:Optimism_vs_Pessimism}
\end{figure}

\begin{figure}
    \centering
    \includesvg[width=0.99\columnwidth,inkscapelatex=false]{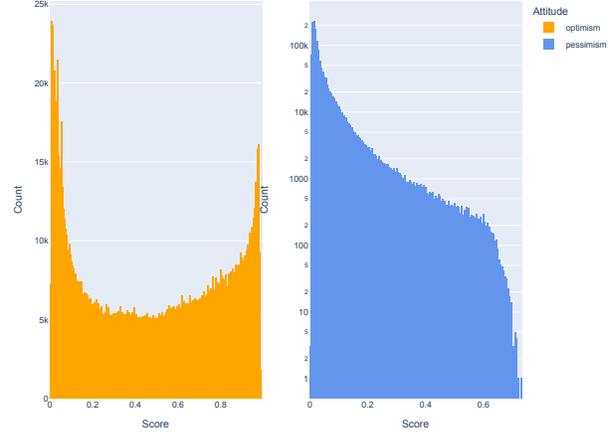}
    \caption{Distribution of Optimism and Pessimism Across the Corpus}
    \label{fig:density_optimism_pessism}
\end{figure}

\begin{figure}
    \centering
\includesvg[width=1.0\linewidth,inkscapelatex=false]{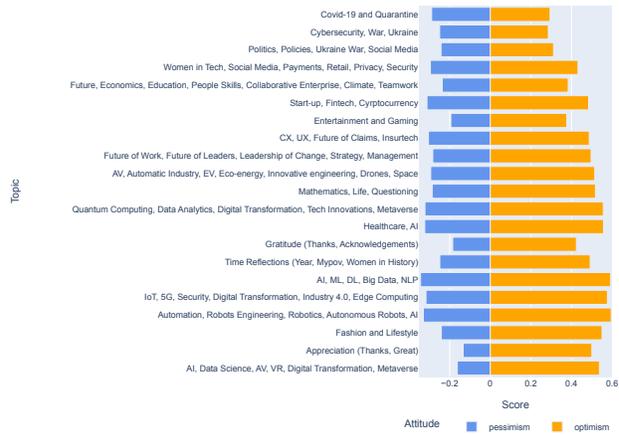}
\caption{Pessimism and Optimism Across Topic Clusters}
\label{fig:optimisim_pessimism_clusters}
\end{figure}

\textbf{Optimistic and Pessimistic Attitudes.} At level of Topics, the median 'Optimism' score (0.9979) is approximately 71.15\% higher than the median 'Pessimism' score (0.5830) (please refer to \Cref{fig:Optimism_vs_Pessimism}). The highest 'Optimism' scores were found in topics, e.g. 35, 43, 73, 53, 65, topics with the lowest 'Optimism' scores were, e.g., 0, 8, 10, 4. The highest 'Pessimism' scores were observed in topics, e.g. 20, 54, 81, 24, 97, 10, whereas the lowest scores included, e.g. 8, 23, 41, 39, 95. A bimodal distribution of 'Optimism' is observed (please refer to \Cref{fig:density_optimism_pessism}). Non-tech topics like Topics 51 and 94 show the highest 'Optimism,' while tech topics likes 79 or 81 have a narrower range of 'Optimism' scores. At level of Clusters, the median 'Optimism' score (0.4963026) is approximately 73.29\% higher than the median of 'Pessimism' score (0.2864076). Clusters demonstrating the highest 'Optimism' scores are Clusters 7, 1, and 5, while the highest 'Pessimism' score are Clusters 1, 7, and 5 (please refer to \Cref{fig:optimisim_pessimism_clusters}).  


\section{Discussion}

\subsection{Anticipated 'Futures-in-the-Making'} 

Among tech-driven discussions, topics related to 'Disruptive TV/Video,' 'Climate Change,' 'War in Ukraine,' 'ML/Data Science/DL/AI,' and 'IoT' generate the most dynamic engagement. In contrast, the topics of 'Robotics, Robots,' 'COVID-19, quarantine, things to do,' 'Healthcare, AI,' 'Innovation engineering, energy technology,' and 'Manufacturing, Agtech, sustainability, innovation' are discussed less frequently. All of them are open-ended 'future present’ \citep{luhmann_differentiation_1982}, still evolving \citep{adam_futures_2011}.

Major events trigger topic spikes, amplifying tech dialogues globally. These peaks can guide tech progress and shape discourse \citep{allcott_social_2017, falkenberg_growing_2022, hopke_visualizing_2018, liang_dynamics_2023}. Inter-topic dynamics, like the alignment in 'AI,' 'VR/AR,' and 'Big Data,' highlight the interconnectedness of tech advancements. These topics clearly stimulate developments in others, revealing a complex network of technological interrelations \citep{porter_tech_2005}. AI serves as a prime example of this phenomenon.

\subsection{Influence of KOLs in Shaping Anticipatory Discourse}
KOLs like Jamesvgingerich, digital\_trans4m, and kirkdborne shape discourse in fields such as 'Climate Change' and 'Robotics.' They foster communities and gain recognition in distinct subjects \citep{furini_x_2024, haupt_facebook_2021, lichti_decentralized_2023}. They not only establish an influential ‘anticipatory news infrastructure’ \citep{ananny_anticipatory_2020}, but also enhance their tweets and foster societal approval for new technologies by integrating anticipatory emotions into the formation of futures \citep{valente_identifying_2007}. The discourse power concentration among few influencers highlights their central role in content creation and dissemination \citep{martin_getting_2022, oueslati_recognition_2023}. A few voices significantly shape public perceptions of technological futures, what raises questions about the diversity and inclusivity of the tech discourse. 

\subsection{Positive Sentiment, Hope, and Optimism Towards Anticipated Futures}

Most of the discourse remained neutral, indicating that influencers tended to provide informative yet potentially ambiguous information (e.g., ‘Machine Learning,’ 'Deep Learning,' 'Climate Change,' 'Disruptive TV/Video,' 'AI,' ‘Future of Claims,’ ‘Digital Transformation,’ and ‘IoT'), possibly due to professional reasons. The prevailing positive sentiment favours the benefits, often without fully addressing the accompanying challenges of technological progress. Conversely, negative sentiments primarily centred on non-technology subjects also discussed by KOLs, such as politics (e.g., 'War in Ukraine', 'Trump Politics'), or social issues resuting from technology (e.g.'Privacy','Cybersecurity'). 

In KOLs’ tweets, a high presence of 'Anticipation' as an emotion (48\%) exceeds the English tweet average (13.9\%) \citep{mohammad_semeval-2018_2018}. 'Anticipation,' linked to future positive events \citep{tenhouten_emotions_2023}, and 'Joy,' as a response to favourable stimuli, are prevalent in the corpus \citep{emmons_joy_2020}. 'Fear,' 'Disgust,' and 'Anger' often coexist, indicating discussions touch on moral or ethical issues \citep{cannon_james-lange_1927, zhan_distinctive_2015, russell_bodily_2013}. For example, war, politics, and data privacy discussions trigger 'Disgust' and 'Anger,' while 'Cybersecurity' debates elicit 'Anger' and 'Fear.' These emotions, especially in morally charged situations, have distinct triggers and outcomes, and can predict moral outrage \citep{hutcherson_moral_2011, salerno_interactive_2013}. They play a key role in decision-making and behaviour \citep{butz_anticipatory_2007, butz_behavioral_2017, feil_anticipatory_2022, hoffmann_anticipatory_2003, odou_how_2020}, influencing choices and actions based on future expectations, leading to optimism and motivation or caution.

Referring strictly to anticipatory emotions, 'Hope' prevails over 'Anxiety' throughout the corpus. Topics with the highest 'Hope' scores include 'Leadership of Change,' 'Change Management,' 'CIO,' 'Business,' 'Future of Work,' 'Agile,' 'Customer,' 'CX,' 'Customer Experience,' 'Employees,' 'Manufacturing,' 'Agtech,' 'Sustainability,' 'Innovation,' and 'Leadership, Leaders, Kate Nasser.' 'Trust' fosters 'Hope' \citep{pleeging_characterizing_2022, pleeging_relations_2021}. Hope is inherently tied to anticipation, fostering a forward-looking perspective that helps individuals manage present challenges by focusing on potential future successes \citep{tenhouten_emotions_2023}. This combination can drive proactive behaviour and resilience in the face of adversity \citep{pleeging_relations_2021}. Conversely, lower 'Hope' scores in topics such as 'People Skills Chat,' 'People, Work RT,' 'Climate Change,' 'Education, Future Economy,' 'Remote Work, Future of Work,' and 'Digital, Transformation, Business, Future' suggest that these areas are perceived with more uncertainty or challenges. This reflects concerns about current issues and future outcomes, highlighting the perceived difficulties and risks associated with these topics. For instance, the ambiguity surrounding climate change's economic  consequences contributes to a pervasive sense of uncertainty \citep{cerreia-vioglio_making_2022}. Additionally, economic policy uncertainty related to environmental regulations can lower hope and increase anxiety about future outcomes \citep{barnett_climate_2022}.

Topics with the highest 'Anxiety' scores include 'CIO, Business, Future of Work, Agile,' 'Manufacturing, Agtech, Sustainability, Innovation,' 'CX, Customer, Customer Experience, Experience,' 'Quantum Computing,' and 'Innovation, Engineering, Energy Technology.' In contrast, topics with the lowest 'Anxiety' scores encompass 'Leadership of Change, Change Management,' 'People Skills Chat, People, Work RT,' 'Customer, CX, Customer Experience, Employees,' 'Leadership, Leaders, Kate Nasser, Future of Work,' and 'Climate Change.' The expectations related to these topics, as set by KOLs, introduce 'Fear,' which is a response triggered by the appraisal of imminent danger \citep{kurth_moral_2015, kurth_anxious_2018}. This 'Fear' can subsequently activate preparatory behaviours for an anticipated future \citep{castelfranchi_anticipation_2011}. Additionally, it is linked with 'Anticipation' (together inducing ‘Anxiety’) related to the uncertainty surrounding certain threats \citep{kurth_moral_2015, kurth_anxious_2018}, particularly the ambiguity regarding their occurrence \citep{liu_neural_2024,vazard_feeling_2024}.

Dominant KOLs' 'Optimism' might impact followers’ views on tech’s potential, reflecting a trend towards techno-optimism \citep{konigs_what_2022}. Although there is some 'Pessimism' that prompts reflection on technology’s potential drawbacks—like privacy concerns or job displacement—this tends to be minimal.

\section{Conclusion}

Incorporating topic modelling, sentiment, and emotion analyses into the context of discourse analysis of 1.5 million tweets allowed us to discover anticipatory discussions on technological futures introduced by KOLs to a public audience. The most talked-about topics on X were 'ML/Data Science/DL/AI,' 'Disruptive TV/Video,' 'IoT,' 'Climate Change,' and 'War in Ukraine.' The results underscore the profound influence of KOLs in shaping these discussions. 

Integrated into discourse analysis, we find that technological discourse is inherently anticipatory, with KOLs' setting a hopeful emotional tone. The high presence of 'Anticipation' combined with 'Trust' results in 'Hope,' while combined with 'Fear,' it leads to 'Anxiety.' Clearly, KOLs propagate 'Hope' rather than 'Anxiety.' This often overlooks potential future threats from technological uncertainty and fosters users' anticipation of a positive technological future. The high presence of 'Anticipation' coupled with 'Joy' leads to 'Optimism,' whereas paired with 'Sadness,' it results in 'Pessimism.' KOLs emphasise 'Optimism' and its associated technological advantages, favouring it over 'Pessimism' and its technological challenges. 

This study contributes to understanding the meaning of anticipatory discourse and its use in influencing public expectations and behaviours. It underscores the significant role of KOLs in guiding societal responses to technological advancements and shows how techno-optimism is propagated in public discourse about future technologies. Moreover, this study advances the field methodologically by employing advanced quantitative methods such as state-of-the-art topic modelling, sentiment analysis, emotion analysis, and attitude analysis. These methods provide a more nuanced and contextually aware analysis of large text corpora, enhancing the precision and scalability of topic modelling in social media discourse.

This study carries at least two substantial theoretical and practical implications. Theoretically, the analysis of emotion correlations suggests that anticipation enhances positive feelings (like trust) while also potentially introducing mild concerns (like fear). This indicates that anticipation and trust together can reliably measure hope. Secondly, the relationship between anticipation and fear, although weaker, can still be used to measure anxiety. The moderate negative correlation between anticipation and sadness supports using anticipation and sadness together to measure pessimism, as increased anticipation is associated with decreased sadness. Finally, given the very weak correlation between anticipation and joy, this combination may not be as effective for measuring optimism. It is worth considering other combinations or additional factors to measure optimism more accurately. From a practical standpoint, this study underscores the potential of integrating anticipatory discourse into AI simulations. Doing so enhances our ability to prepare for and navigate future uncertainties, making it a critical tool for various fields striving to stay ahead of the curve. This method can complement traditional forecasting and scenario planning techniques, offering a more comprehensive toolkit for anticipating future developments.

It is important to acknowledge limitations of our findings. Firstly, the analysis was limited to a selected group of technology influencers, which may not capture the full spectrum of opinions on future technologies, thus affecting the generalisability of our conclusions. Additionally, we focused solely on tweets, excluding replies and comments that could offer deeper insights into public perceptions. Secondly, our keyword analysis disregarded context and domain, factors that significantly influence word meanings and precision. Thirdly, while this study provides valuable insights into the correlations between emotions, the weak correlation between anticipation and joy suggests that this combination may not be effective for measuring optimism. Future research should explore other combinations or additional factors to measure optimism more accurately. Additionally, the relationship between anticipation and fear, though usable for measuring anxiety, is relatively weak and may benefit from further investigation. Future research should include comparative studies with other media sources, conduct longitudinal studies to track discourse evolution, and refine quantitative methods by incorporating additional data and developing more sophisticated models to capture nuances in anticipatory discourse.


\bibliographystyle{unsrtnat}
\setcitestyle{authoryear,open={((},close={))}} 


\listoftables
\listoffigures

\section{Declarations}

\paragraph{Data availability}
The datasets generated during and/or analysed during the current study are available in the OSF repository, and can be accessed via the following link: \url{https://osf.io/z925y}.

\paragraph{Competing Interests}
The authors declare no potential conflict of interests.

\paragraph{Ethical approval}
Ethical approval was not required for this study.

\paragraph{Informed Consent}
This article does not contain any studies with human participants performed by any of the authors.

\paragraph{Author contributions}
The manuscript was collaboratively drafted by all authors, who also conducted critical revisions and editing. The first author conceptualised the analysis and designed the methods. The second author collected data and performed BERTopic analysis, while the third author focused on sentiment, emotions, and attitude analyses. All authors approved the version to be published.

\end{document}